\pdfoutput=1
\documentclass[11pt]{article}

\usepackage{acl}

\usepackage{times}
\usepackage{latexsym}

\usepackage[T1]{fontenc}
\usepackage[utf8]{inputenc}

\usepackage{microtype}

\usepackage{mathtools}
\usepackage{amsmath}
\usepackage{multirow}

\usepackage{tikz}
\usepackage{tikz-qtree}
\usepackage{tikz-dependency}

\usepackage{xurl}

\title{Auxiliary Tasks to Boost Biaffine Semantic Dependency Parsing}

\author{Marie Candito \\
  LLF, Université Paris Cité - CNRS, Paris, France \\
  \texttt{marie.candito@u-paris.fr} \\
  }

\begin{document}
\maketitle
\begin{abstract}
The biaffine parser of \citet{dozat-biaffine-attention-iclr-17} was successfully extended to semantic dependency parsing (SDP) \cite{dozat-manning-2018-simpler}. Its performance on graphs is surprisingly high given that, without the constraint of producing a tree, all arcs for a given sentence are predicted independently from each other (modulo a shared representation of tokens).
To circumvent such an independence of decision, while retaining the $O(n^2)$ complexity and highly parallelizable architecture, we propose to use simple auxiliary tasks that introduce some form of interdependence between arcs. Experiments on the three English acyclic datasets of SemEval 2015 task 18 \cite{oepen-etal-2015-semeval}, and on French deep syntactic cyclic graphs \cite{ribeyre-tlt13} show modest but systematic performance gains on a near state-of-the-art  baseline using transformer-based contextualized representations. This provides a simple and robust method to boost SDP performance.

\end{abstract}

\section{Introduction and related work}
\begin{figure}
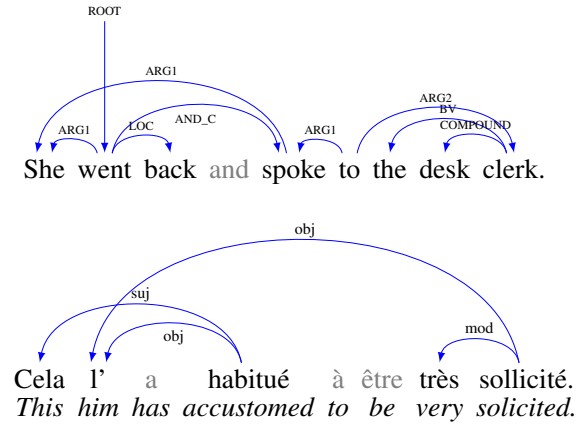

\begin{center}
  \scalebox{0.95}{
    \begin{dependency}[theme=simple,edge horizontal padding=3, edge style={>=latex,blue}]
      \begin{deptext}
      She \& went \& back \& \textcolor{gray}{and} \& spoke \& to \& the \& desk \& clerk.\\
      \end{deptext}
      \depedge[edge end x offset=3pt]{2}{1}{ARG1}
      \depedge[edge unit distance=2ex,edge end x offset=-3pt]{5}{1}{ARG1}
      \deproot[]{2}{ROOT}
      \depedge[]{2}{3}{LOC}
      \depedge[edge unit distance=2ex,label style={below},edge end x offset=-6pt]{2}{5}{AND\_C}
      \depedge[edge end x offset=2pt]{6}{5}{ARG1}
      \depedge[edge unit distance=5ex]{9}{7}{BV}
      \depedge[]{9}{8}{COMPOUND}
      \depedge[edge unit distance=5ex]{6}{9}{ARG2}
      
    \end{dependency}
  }\\
\scalebox{0.95}{
    \begin{dependency}[theme=simple,edge horizontal padding=3, edge style={>=latex,blue}]
      \begin{deptext}
        Cela \& l' \& \textcolor{gray}{a} \& habitué \&  \textcolor{gray}{à} \&  \textcolor{gray}{être} \& très \& sollicité.\\
        \textit{This}  \& \textit{him} \& \textit{has} \& \textit{accustomed} \& \textit{to} \& \textit{be} \& \textit{very} \& \textit{solicited.}\\
      \end{deptext}
      \depedge[edge unit distance=2ex]{4}{1}{\footnotesize{suj}}
      \depedge[edge unit distance=1.1ex,label style={below},edge end x offset=3pt]{4}{2}{\footnotesize{obj}}
      \depedge[]{8}{7}{\footnotesize{mod}}
      \depedge[blue,edge unit distance=0.4ex,edge end x offset=-3pt]{8}{2}{\footnotesize{obj}}
    \end{dependency}
  }
  \caption{\textbf{Top:} English Semantic graph in the DM format, as part of the SemEval2015-Task18 dataset \cite{oepen-etal-2015-semeval}. \textbf{Bottom:} French Deep syntactic graph as defined by \citet{candito-etal-2014-deep}. }
  \label{fig-graphs}
  \end{center}
\end{figure}

Semantic dependency parsing is the task of producing a dependency graph for a sentence. Depending on the datasets, these dependencies may correspond to predicate-argument relations, with labels numbering semantic arguments (as in Figure \ref{fig-graphs}-top) or dependencies with intermediate status between syntax and semantics, with labels being canonical grammatical functions that normalize syntactic alternations (e.g. in Figure \ref{fig-graphs}-bottom, the clitic \textit{l' (him)} is the canonical object of the passive verb form \textit{sollicité (solicited)}).

If one views each dependency as a decision to make, both dependency parsing (DP, outputing a syntactic tree) and semantic dependency parsing (SDP) are known to exhibit high interdependence of decisions. For instance in DP, when parsing \textit{a question answering machine}, choosing \textit{machine} as root is linguistically coherent with the \textit{machine} $\rightarrow$ \textit{answering} $\rightarrow$ \textit{question} analysis only, whereas (wrongly) choosing \textit{question} as root is syntactically coherent with the \textit{question} $\rightarrow$ \textit{answering} $\rightarrow$ \textit{machine} analysis. 

In DP though, the interdependence between arcs is partially solved by the tree constraint: choosing one head for a given token amounts to ruling out all other heads. This structural interdependence is absent in SDP. Complex structural, lexical and semantic factors control whether a given dependent should be attached to zero, one or several heads.

Several approaches exist in the literature to capture interdependence of arcs in SDP, which often derive from proposals made for DP. One is to use a higher-order graph-based parser. \citet{wang-etal-2019-second} achieve state-of-the-art results (without pretrained LM) on the English part of the SemEval2015-Task18 data, by using second-order factors to score the graphs, yet at the cost of a $O(n^3)$ complexity.

Another main approach is to use sequential decisions, and hence take advantage of previous decisions by encoding the previously predicted arcs. 
This is the case in the transition-based parser of \citet{fernandez-gonzalez-gomez-rodriguez-2020-transition}, or in the system of \citet{kurita-sogaard-2019-multi}, which selects a new head for certain tokens at each iteration, using reinforcement learning to 
order this selection of heads. Both models have a $O(n^2)$ complexity (when used without cycle detection), and in both cases, sequential decisions benefit from the encoding of previously predicted arcs, yet at the cost of error propagation. For that reason, \citet{bernard-2021-multiple} propose a system close to that of \citet{kurita-sogaard-2019-multi}, yet allowing the system to overwrite previous decisions and hopefully correct itself.

On the contrary, the biaffine system of \citet{dozat-manning-2018-simpler} (hereafter {\bf DM18}) performs a simultaneous scoring of all candidate arcs, decides to predict an arc independently of the other ones. This results in a highly parallelizable $O(n^2)$ inference, with surprisingly high performance albeit below second-order parsing. 

As for most NLP tasks, SDP performance increases when integrating transformer-based contextual representations when encoding input tokens. On the English dataset from the SemEval 2015 Task 18 \cite{oepen-etal-2015-semeval},  
\citet{fernandez-gonzalez-gomez-rodriguez-2020-transition} (hereafter {\bf FG20}) report a +0.7 and +2.0 increase for the in-domain (ID) and out-of-domain (OOD) test sets respectively.\footnote{\label{foot-hc}Using the biaffine DM18 architecture, \citet{he-choi-FLAIRS-2020} report a +2 and +3 point increase in ID and OOD. Yet, these results are not comparable: the authors have used a different pre-processing, which adds orphan dependencies from the root to orphan tokens, resulting in an easier task (p.c. with the authors and\\ \url{https://github.com/emorynlp/bert-2019/issues/1}).}

In this work, we retain the simple $O(n^2)$ biaffine architecture of DM18, and we investigate how simple auxiliary tasks can introduce some interdependence between arc decisions, in a multi-task learning setting \citep{caruana-1997}. We show modest but statistically significant improvements on the three English datasets of the widely used SemEval2015-Task18 data \citep{oepen-etal-2015-semeval}. We also test another appealing property of the biaffine architecture, which is the absence of formal constraints on the output graphs. Experiments on French deep syntactic graphs \citep{ribeyre-tlt13}, which are highly cyclic, also demonstrate the effectiveness of our auxiliary tasks for SDP.

\section{The baseline biafine graph parser}

We reuse the computation of the arc and label scores of the DM18 model, which we modernized by using contextual representations: input sequence $w_{1:n}$ is passed into a pretrained language model. We represent a word-token $w_i$ by concatening the contextual vector of its first subword\footnote{\citet{he-choi-FLAIRS-2020} report very slight improvements in OOD when using the average of all subwords, but an opposite trend in ID.} $\mathbf{h}_{i}^{(\text{bert})}$ and a word embedding $\mathbf{e}_{i}^{(\text {word})}$.

\begin{equation}
\begin{array}{l}
\mathbf{v}_{i} =\mathbf{h}_{i}^{(\text{bert})} \oplus \mathbf{e}_{i}^{(\text {word})}\\
\end{array}
\label{eq-vi-one}
\end{equation}

For some of the experiments, we also concatenate a lemma and a POS embedding. % \footnote{In all our experiments, we use the hidden BERT representation at the last layer, of the first subword of a word.}
\begin{equation}
\begin{array}{l}
\mathbf{v}_{i} =\mathbf{h}_{i}^{(\text{bert})} \oplus \mathbf{e}_{i}^{(\text {word})} \oplus \mathbf{e}_{i}^{(\text {lemma})} \oplus \mathbf{e}_{i}^{(\text {POS})} \\
\end{array}
\label{eq-vi-two}
\end{equation}

The sequence of word-tokens representations is passed into several biLSTM layers: $\mathbf{r}_{1:n} =\operatorname{biLSTM}(\mathbf{v}_{1:n})$.

The recurrent representation $r_i$ is then specialized according to two binary features: head versus dependent, and arc versus label score:
\vspace{-1mm}
\begin{equation}
\begin{array}{l}
\mathbf{h}_{i}^{(\text {arc-head})}=\mathrm{MLP}^{(\text {arc-head})}\left(\mathbf{r}_{i}\right) \\
\mathbf{h}_{i}^{(\text {lab-head})}=\mathrm{MLP}^{(\text {lab-head})}\left(\mathbf{r}_{i}\right) \\
\mathbf{h}_{i}^{(\text {arc-dep})}=\mathrm{MLP}^{(\text {arc-dep})}\left(\mathbf{r}_{i}\right) \\
\mathbf{h}_{i}^{(\text {lab-dep})}=\mathrm{MLP}^{(\text {lab-dep})}\left(\mathbf{r}_{i}\right) \\
\end{array}
\end{equation}

We use a simplified biaffine transformation for arc scores, and a per-label one for label scores:
\begin{equation}
\begin{array}{l}

s_{i \rightarrow j}^{(\text {arc})} = \mathbf{h}_{j}^{(\text {arc-dep})} \mathbf{U}^{(\text {arc})} \mathbf{h}_{i}^{(\text {arc-head})\top}+\mathbf{b}^{(\text {arc})} \\

s_{i \rightarrow j}^{(\text {l})} = \mathbf{h}_{j}^{(\text {lab-dep})} \mathbf{U}^{(\text {l})} \mathbf{h}_{i}^{(\text {lab-head})\top}+\mathbf{b}^{(\text {l})}

\end{array}
\label{eq-arc-label-scores}
\end{equation}

For each position pair $i,j$, a binary cross-entropy loss is used for the existence of arc $i\rightarrow j$, and a cross-entropy loss is used for the labels of gold arcs. At inference time, any candidate arc with positive score $s_{i \rightarrow j}^{(\text {arc})}$ is predicted, and receives the label with maximum score for this arc.

\section{Auxiliary tasks targeting sets of arcs}

\begin{figure}
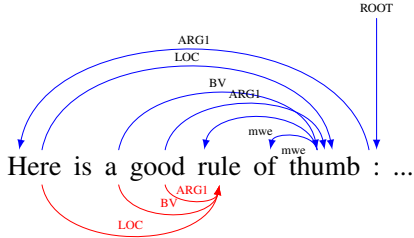

\begin{center}
  \scalebox{0.95}{
    \begin{dependency}[theme=simple,edge horizontal padding=3, edge style={>=latex,blue}]
      \begin{deptext}
      Here \& is \& a \& good \& rule \& of \& thumb \& : \& ...\\
      \end{deptext}
      \depedge[edge end x offset=3pt]{1}{7}{LOC}
      \depedge[edge unit distance=2ex,edge end x offset=-3pt]{3}{7}{BV}

      \depedge[]{4}{7}{ARG1}
      \depedge[edge unit distance=2ex,label style={below},edge end x offset=-6pt]{7}{5}{mwe}
      \depedge[edge end x offset=2pt,label style={below}]{7}{6}{mwe}
      \depedge[edge below, edge style={red}]{4}{5}{\textcolor{red}{ARG1}}
      \depedge[edge below, edge style={red}]{3}{5}{\textcolor{red}{BV}}
      \depedge[edge below, edge style={red}]{1}{5}{\textcolor{red}{LOC}}
      \deproot[]{8}{ROOT}
      \depedge[edge end x offset=-6pt]{8}{1}{ARG1}
    \end{dependency}
  }
  \caption{Example of competition for the sequence \textit{rule of thumb}. Above arcs: correct MWE analysis (\textit{rule} and \textit{of} attached to the last MWE component \textit{thumb}, and \textit{thumb} being the head of the sequence). Below arcs: incorrect compositional analysis, in which \textit{rule} is the head, e.g. attached wrongly as ARG1 of \textit{good} (in red).}
  \label{fig-incorrect}
  \end{center}
\end{figure}

Preliminary experiments on English semantic dependency graphs \citep{oepen-etal-2015-semeval} and on French deep syntactic graphs \citep{candito-etal-2014-deep} have shown us that the biaffine graph parser gives good results, but with inconsistencies that are clearly related to the locality of decisions. In particular, a quick error analysis revealed incompatible arc combinations. More precisely, we noticed impossible sets of labels for the set of heads of a given dependent. For example in the DM part of the SemEval2015-Task18 dataset, tokens are sometimes attached with a \texttt{mwe} label (for a component of a multi-word expression) and attached to another head with a non-mwe label, as shown for the \textit{rule} token in Figure \ref{fig-incorrect}. This incorrect situation actually never happens in the training set, but this impossibility is not captured by the model. In the predicted French deep syntactic graphs, we noticed punctuation tokens wrongly attached to two different heads with the specific \texttt{punct} dependency label. 

A second observation is a tendency in some of the datasets to predict disconnected tokens (i.e. with no incoming nor outgoing arcs) too frequently. More generally, when counting the number of predicted heads for each token in the predicted graphs, the accuracy is about 95\% in the English datasets, and below 92\% for the French one.

\subsection{Auxiliary tasks}

Hence the idea of using auxiliary tasks taking into account all the heads (resp. dependents) of a given token. More precisely, we experiment multi-task learning on the two target tasks (tasks {\bf A} and {\bf L} for arc and label prediction), plus the following auxiliary tasks, which predict for each token $w_j$:
\begin{itemize}
\item tasks {\bf H} and {\bf D}: the number of governors and number of dependents. For instance in top Figure \ref{fig-graphs}, \textit{spoke} has two governors (\textit{went} and \textit{to}) and one dependent (\textit{She});
\item the labels of the incoming arcs, either as:
\begin{itemize}
    \item task {\bf S}: the concatenated string of the incoming arcs labels, in alphabetic order (e.g. for the \textit{spoke} token, \texttt{AND\_C+ARG1})
    \item task {\bf B}: or the "bag of labels" (BOL) sparse vector, whose components are the numbers of incoming arcs to $w_j$ bearing each label. For the \textit{spoke} token, this would give a 1 for the \texttt{AND\_C} label component and 1 for the \texttt{ARG1} label, and zeros for all other labels. 
\end{itemize}
\end{itemize}

Technically, for each auxiliary task, a specific MLP is used to specialize the recurrent representation $r_j$ of each word-token $w_j$.
 Tasks H and D are regression tasks, which use MLPs with a single output neuron and a squared error loss.\footnote{$nbh_j$ is interpreted as $1+$ the log of the number of heads of $w_j$, and same for $nbd_j$, so as to penalize less errors in bigger numbers: e.g penalize more predicting one head instead of 0 than predicting 2 heads instead of 3.} 

{\begin{center}
$
\begin{array}{l}
  nbh_j = \mathrm{MLP}^{(\text{H})}(\mathbf{r}_j)  \\
  nbd_j = \mathrm{MLP}^{(\text{D})}(\mathbf{r}_j)  \\
\end{array}
$
\end{center}}

The S task is a classification task into categories corresponding to multi-sets of labels encountered in the training set.\footnote{The categories are label multi-sets because we neutralize the order of the heads when considering the incoming arcs. This limits the number of categories.} For $w_j$, the vector of scores of all the label multi-sets is $\textbf{s}_{j} = \mathrm{MLP}^{(\text{S})}(\mathbf{r}_j)$, and cross-entropy loss is used at training.

For the B task, we use a MLP with final layer of size $|L|$: $ \textbf{BOL}_j = \mathrm{MLP}^{(\text{B})}(\mathbf{r}_j)$. The component for label $l$, $ \text{BOL}_{jl}$, is interpreted as $1+$ log of the number of 
$l$-labeled incoming arcs to $w_j$. 
The loss we use is the L2 distance between gold and predicted BOL vectors.\\

The two example inconsistencies cited above are indirectly captured by these auxiliary tasks. Firstly, in case of a token $w_j$ that is a component of a multi-word expression, the recurrent representation $\mathbf{r}_j$ for this token will be optimized to lead to a single \texttt{mwe} incoming label for the S or B tasks, and a value 1 for the H task (cf. a single governor for $w_j$). Hopefully, when used for the A and L tasks, $\mathbf{r}_j$ will favor incoming arcs from close next tokens (cf. e.g. for the DM format, mwe components are attached to the right, and quite locally). The other mentioned problem of predicting disconnected tokens too frequently is captured by the H and D tasks. Predicting no incoming nor outgoing arc for a given token will only be coherent with H=0 and D=0 for this token. Hence, hopefully, predicting more than 0 for the H task for a token $w_j$, will lead to higher scores for arcs pointing to $w_j$.

\subsection{Combining sublosses}
\label{sec-loss}
At training time, for each batch, we seek to minimize a weighted sum of the losses for all the tasks, whether main or auxiliary. Manually tuning these weights is cumbersome and suboptimal. We use the notion of task uncertainty and the approximation proposed by \citet{kendall-uncertainty-weight-loss-CVPR-2018}, who introduce a parameter $\sigma_t$ for each subtask $t$, to be interpreted as its ``uncertainty''. Noting $T$ as the set of tasks, the overall loss for a batch is
$
\sum_{t \in T} \frac{1}{\sigma_t^2}L_t + \text{ln}(\sigma_t)
$.

The parameters $\sigma_t$ are initialized to 1 and modified during the learning process. The first term of the sum ensures that the more uncertain the task, the less its loss will count, while the second term prevents arbitrarily augmenting the $\sigma_t$ values, thus reducing the loss weights.

\subsection{Stack propagation}
We test two multi-task learning configurations: first, simple parameter sharing up to the biLSTM layers (all specialization MLPs applied on the recurrent token representations). Second, we test the technique of ``\textit{stack propagation}'', which \citet{zhang-weiss-2016-stack} experimented for the POS tagging and parsing tasks. In our case, it amounts to using the dense layers of the auxiliary tasks MLPs to score the arcs and their labels.

For example, to use the H task in stack propagation mode, let $\text{hidden}_j^{(\text{H})}$ be the hidden layer of $\mathrm{MLP}^{(\text{H})}$ for the dependent $j$, and $c^{(\text{H})}$ a coefficient hyperparameter. The computation of $s_{i \rightarrow j}^{(\text {arc})}$ (cf. equation \ref{eq-arc-label-scores}) is modified as follows:
{\begin{center}
$
\begin{array}{l}

\mathbf{sp}_j^{(\text {arc-dep})} = \mathbf{h}_{j}^{(\text {arc-dep})} \oplus c^{(\text{H})} \text{hidden}_j^{(\text{H})}\\
s_{i \rightarrow j}^{(\text {arc})} = \mathbf{sp}_j^{(\text {arc-dep})} \mathbf{U}^{(\text {arc})} \mathbf{h}_{i}^{(\text {arc-head})\top}+\mathbf{b}^{(\text {arc})} \\
\end{array}
$
\end{center}}

Similarly, to use task B in stack propagation mode, we modify the score of each label:

{\begin{center}
$
\begin{array}{l}

\mathbf{sp}_j^{(\text {lab-dep})} = \mathbf{h}_{j}^{(\text {lab-dep})} \oplus c^{(\text{B})} \text{hidden}_j^{(\text{B})}\\

s_{i \rightarrow j}^{(\text {l})} = \mathbf{sp}_{j}^{(\text {lab-dep})} \mathbf{U}^{(\text {l})} \mathbf{h}_{i}^{(\text {lab-head})\top}+\mathbf{b}^{(\text {l})}

\end{array}
$
\end{center}}

Note that it 
forces to perform the auxiliary tasks during inference, instead of at training time only.

\section{Experiments and discussion}

\subsection{Datasets}
We experiment on the three widely used English datasets of SemEval2015-Task18 \citep{oepen-etal-2015-semeval} (DM, PAS and PSD), which are acyclic graphs mainly representing predicate-argument relations. We also experiment on French deep syntactic graphs  \citep{ribeyre-tlt13} (Appendix \ref{sec-full-res-french}). These capture most of argument sharings (e.g. raising, obligatory and arbitrary control, subject sharing in VP coordination) but are closer to surface syntax in the sense that labels remain syntactic, even though syntactic alternations are neutralized (e.g. passive {\textit{by}}-phrases are labeled as subjects). Cycles may appear e.g. in relative clauses.\footnote{In these deep graphs, the root tokens (usually unique) are attached to a dummy root token in practise. Thus for this dataset, in all the above formulations, the sequence $w_{1:n}$ corresponds to a sentence of $n-1$ word-tokens, with a dummy root $w_1$. For its contextual representation, we use the contextual vector of the beginning of sequence token.}

\subsection{Experimental protocol}

We chose to investigate the impact of the auxiliary tasks on a high baseline, using pretrained contextual representations.
We use our own implementation\footnote{\footnotesize{\url{https://github.com/mcandito/aux-tasks-biaffine-graph-parser-FindingsACL22}}} of the biaffine parser, the BERT$_{\text{base-uncased}}$  model for English, and FlauBERT$_{\text{base-cased}}$ for French.\footnote{We used the HuggingFace library \citep{wolf-etal-2020-transformers}.} We used two settings (see Appendix \ref{sec-appendix-hyper} for details): 
\begin{itemize}
    \item {\bf BERT$_{\text{tuned}}$}: the first setting is intended to use the contextual representations as only source of pre-trained parameters, and defines the $\mathbf{v}_i$ vectors as in equation (\ref{eq-vi-one}) (no lemma nor POS embeddings), with the word embeddings being randomly initialized, and the BERT embeddings being fine-tuned for the SDP task. 
    \item {\bf BERT$_\text{froz}$+POS+lem}: the second one is used to compare our results to previous work on the English SemEval2015-Task18 datasets: the BERT embeddings are frozen, additional POS and lemma embeddings are used (cf. $\mathbf{v}_i$ definition as in equation (\ref{eq-vi-two})). The same pre-trained word and lemma embeddings as FG20 are used. Note this setting uses gold POS and lemmas and is not a realistic scenario.
\end{itemize}
For the {\bf BERT$_{\text{tuned}}$} mode, we tuned the hyperparameters on the French data, and applied the same configuration to the English datasets. After a few tests, we set a configuration (see Appendix \ref{sec-appendix-hyper}), and searched for the best combination of auxiliary tasks. For each experiment, we report the labeled Fscore (LF), including root arcs, averaged over 9 runs.

\subsection{Results on French deep syntactic graphs}

Table \ref{tab-res-combis-taches-aux} shows the results with and without various combinations of auxiliary tasks.\footnote{Previous state-of-the art on this data is a non-neural system: \citet{ribeyre-etal-2016-accurate} obtained LF=80.86, and went up to LF=84.91 thanks to features from constituency parses from the rich FrMG parser \citep{villemonte-de-la-clergerie-2010-convertir}. The biaffine architecture with contextual vectors, without auxiliary tasks, obtains a mean LF=86.79.} While no auxiliary task provide a significative increase\footnote{See Appendix \ref{sec-full-res-french} for results with each auxiliary task.}, the combinations B+H and B+D+H+S bring a statistically significant $+0.53$ and $+0.56$ increase on average (see Appendix \ref{sec-significance-test} for significance test).

\begin{table}[h]
\begin{center}
  \begin{tabular}{|l|l|l|c|}
\hline

Auxiliary  	& Stack & \multicolumn{2}{|c|}{On 9 runs}\\\cline{3-4}
tasks	&  propagation & meanLF	& stdev\\\hline
$\emptyset$ & NA	& 86.79	& 0.19\\\hline

B+H	& no & 87.32$^{***}$	& 0.18\\ 
D+H	& no & 86.61	& 0.71\\
  H+S	& no & 87.04	& 0.17\\\hline

B+D+H+S	& no & {\bf 87.35}$^{***}$	& 0.26\\\hline 
B+H     & $c^{(\text{B})}$=1 $c^{(\text{H})}$=1	& 87.49	& 0.06\\
B+H     & $c^{(\text{B})}$=1 $c^{(\text{H})}$=10	& {\bf 87.66}$^{\text{+++}}$	& 0.18\\\hline 
  \end{tabular}
  \caption{Results on French dev set, for various tasks combinations, with and without stack propagation (H: nb of heads, D: nb of dependents, B: bag of labels, S: label multi-set). Col3-4: average LF, and standard deviation. $^{***}$: significative diff. wrt first line. $^{\text{+++}}$: significative diff. wrt second line ($p <$0.001).}
  \label{tab-res-combis-taches-aux}
\end{center}
\end{table}

We tested the impact of stack propagation using auxiliary tasks B+H. We observe a modest but statistically significant $+.34$ increment with weights $c^{(\text{B})}$=1 and $c^{(\text{H})}$=10. 
Considering that this makes the inference task more complex, we did not use it in later experiments on English.

\subsection{Results on English semantic graphs}

\begin{table}[h]
\begin{center}
  \begin{tabular}{|c|l|l|l|l|l|}
\hline
  & Tasks & DM & PAS & PSD & Avg \\\hline
  \multirow{2}{*}{ID} & $\emptyset$ & 93.7 & 93.9 & 80.7 & 89.4 \\
                      & B+H         & 94.2 & 94.3 & 81.2 & 89.9 \\\hline
  \multirow{2}{*}{OOD} & $\emptyset$ & 90.3 & 92.0 & 79.8 & 87.4 \\
                      & B+H         & 91.0 & 92.8 & 80.2 & 88.0 \\\hline
  \end{tabular}
  \caption{Average LF (on 9 runs), in BERT$_{\text{tuned}}$ setting, on English in-domain (ID) and out-of-domain (OOD) test sets, using either no auxiliary task ($\emptyset$) or tasks B and H (B+H), without stack propagation. B+H results are statistically higher than $\emptyset$ for DM ID, DM OOD, PAS ID, PAS OOD ($p <$0.001) and PSD ID ($p <$0.01).}
  \label{tab-res-english-test}
\end{center}
\end{table}

We then tested the B+H configuration on the English test sets. In Table \ref{tab-res-english-test}, we observe  that performance gains using B and H auxiliary tasks are systematic across datasets (DM, PAS, PSD) and across in- or out-of-domain test sets, which tends to show the robustness of our method.\footnote{The improvement tends to be higher for OOD, and for DM and PAS. One reason could be that the PSD graphs show less reentrancies, hence the number of heads is more predictable, and using a specific auxiliary task for it is less beneficial.}

We can also measure the impact of the auxiliary tasks by evaluating how accurate the predicted graphs are, concerning the number of heads of tokens: on average on the English dev sets, the proportion of tokens receiving the right number of heads in the predicted graphs increases from $94.9$ without auxiliary tasks to $95.5$ with tasks B+H.

Finally, we provide in Table \ref{tab-comparison-english} a comparison to FG20 results (thus using the BERT$_{\text{froz}}$+POS+lem setting), which are the state-of-the-art for systems using a single source of contextual embeddings. While our results remain below, note that our auxiliary tasks can be used with their system, 
as well as with e.g. that of \citet{wang-etal-2021-automated}, which achieve significant improvements with an automated concatenation of various contextual embedding models, reaching 91.7 for ID et 90.2 for OOD. 

\begin{table}[h]
\begin{center}
  \begin{tabular}{|l|c|c|}
\hline
  & ID & OOD \\\hline
  {\bf FG20} BERT$_{\text{froz}}$+POS+lem  & 90.7 & 88.8\\
  {\bf Ours} BERT$_{\text{froz}}$+POS+lem, B+H & 90.2 & 87.9\\\hline
    {\bf Ours} BERT$_{\text{tuned}}$, B+H & 89.9 & 88.0\\\hline
  \end{tabular}
  \caption{Comparison to the state-of-the-art SDP parser using BERT, on English ID and OOD test sets, in {\bf BERT$_{\text{froz}}$+POS+lem} setting. {\bf FG20}: \citep{fernandez-gonzalez-gomez-rodriguez-2020-transition}.}
  \label{tab-comparison-english}
\end{center}
\end{table}

\section{Conclusion}
When using a biaffine graph-based architecture for semantic dependency parsing (SDP), arcs are predicted independently from each other. 
Our contribution is a set of simple yet original auxiliary tasks that introduce some form of interdependence of arc decisions. We showed that training recurrent word-token representations both for the SDP task and for predicting the number of heads and the incoming labels of each word is systematically beneficial, when tested either on English or French, on semantic or on deep syntactic graphs, and on in- or out-of-domain data. 

\bibliography{anthology,custom}
\bibliographystyle{acl_natbib}

\appendix

\section{Training details}
\label{sec-appendix-hyper}
 All experiments are run on a Nvidia GTX 1080 Ti GPU.
 \paragraph{For all settings:}
\begin{itemize}
\item lexical embedding size ($e^{(word)}_i$) : 100
\item lexical dropout : 0.4. At learning, the lexical embedding of a token is replaced with probability 0.4 by a special token *DROP*, whose embedding is learned.
\item biLSTM : 3 layers of size $2 * 600$, with $0.33$ dropout
\item MLPs for aux. tasks: 1 hidden layer (300), output layer (300), dropout $0.25$
\item Batch size = 8
\item Optimizer = Adam, $\beta_1=\beta_2=0.9$
\item Learning is stopped when all labeled Fscores decrease on the dev set. Note that beside the main FL score, tasks H and B give rise to their labeled Fscore, computed using the number of heads as predicted by task H (resp. B).
\end{itemize}

\paragraph{BERT$_{\text{tuned}}$ setting}
\begin{itemize}
\item BERT and FlauBERT models : fine-tuned
\item MLPs for arc and label score : 1 hidden layer (600), output layer (600), dropout $0.33$
\item Learning rate = $2\times 10^5$
\item Loss combination : learnt weights (cf. section \ref{sec-loss})
\end{itemize}

 \paragraph{BERT$_{\text{froz}}$+POS+lem setting}
\begin{itemize}

\item BERT model : frozen
\item word and lemma embeddings (100), initialized with embeddings by \citet{ma-etal-2018-stack}, fine-tuned
\item POS embeddings (100), randomly initialized
\item MLP for arc score : 1 hidden layer (500), output layer (500), dropout $0.33$
\item MLP for label score : 1 hidden layer (100), output layer (100), dropout $0.33$\footnote{Sizes of the MLPs for arc and label scores are defined here to replicate \cite{he-choi-FLAIRS-2020} settings. We observed very marginal differences when keeping the sizes used in BERT$_{\text{tuned}}$ setting (600 for both MLPs).}
\item Learning rate = $5\times 10^4$
\item loss combination : plain sum of losses for each task
\end{itemize}

\section{Unsuccessful tests}
\label{sec-unsuccessful-tests}
Various tests were abandoned as unsuccessful in our preliminary tests:

\begin{itemize}
\item Using pre-trained lexical embeddings with tuned contextual embeddings had no impact on performance on average.
\item Freezing BERT's and FlauBERT's parameters without using word and lemma embeddings significantly decreased performance (by about 2 FL points).
\item Increasing the level of parameter sharing between tasks was not successful: instead of applying the MLPs of the auxiliary tasks on the recurrent representations $\mathbf{r}_i$, we tested applying them on the outputs of the specialization MLPs (i.e. on $\mathbf{h}_{i}^{(\text {arc-head})}$ for task D, on $\mathbf{h}_{i}^{(\text {arc-dep})}$ for task H, on $\mathbf{h}_{i}^{(\text {lab-dep})}$ for tasks B and S). While this tends to increase the number of epochs, it does not improve the performance. 
\end{itemize}

\section{Significance testing}
\label{sec-significance-test}

We use a Fisher-Pitman exact permutation test to estimate  the significance of the differences in performance between two configurations (as done for example by \cite{bernard-2021-multiple}). More precisely, suppose we consider two samples of Fscores, for $nA$ runs corresponding to configuration A, and $nB$ runs for configuration B, with on average configuration B better than A. The null hypothesis is that the two samples follow the same distribution. The p-value corresponds to the probability that separating the set of Fscores into two samples A' and B' of size $nA$ and $nB$ gives a difference in mean at least as large as the observed difference. With the exact test, the p-value is calculated exactly, on all possible splits into A' and B' samples.

\section{French data statistics and full results}
\label{sec-full-res-french}

\begin{table}[h]
\begin{center}
  \begin{tabular}{|r|r|r|}
\hline
  & Train & Dev \\\hline
  Nb of sentences & 14,759 & 1,235\\
  Nb of tokens & 457,872 & 40,055\\
  \% of disconnected tokens & 12.0 & 12.2\\
  Nb of edges & 424,813 & 37,110\\\hline
  \end{tabular}
  \caption{Statistics of the deep French syntactic graphs, built on the French treebank \citep{abeille-barrier-2004-enriching}.}
  \label{tab-french-data-statistics}
\end{center}
\end{table}

The complete results on dev set for the French data is provided in Table \ref{tab-full-res-combis-taches-aux}, of which Table \ref{tab-res-combis-taches-aux} is a truncated version.

\begin{table}[h]
\begin{center}
  \begin{tabular}{|l|l|l|c|}
\hline

Auxiliary  	& Stack & \multicolumn{2}{|c|}{On 9 runs}\\\cline{3-4}
tasks	&  propagation & meanLF	& stdev\\\hline
$\emptyset$ & NA	& 86.79	& 0.19\\\hline

H	& no & 86.82	& 0.54\\
D	& no & 86.83	& 0.40\\
S	& no & 86.98	& 0.30\\
B	& no & 87.05	& 0.49\\\hline

B+H	& no & 87.32$^{***}$	& 0.18\\ 
D+H	& no & 86.61	& 0.71\\
  H+S	& no & 87.04	& 0.17\\\hline

B+D+H+S	& no & {\bf 87.35}$^{***}$	& 0.26\\\hline 
B+H     & $c^{(\text{B})}$=1 $c^{(\text{H})}$=1	& 87.49	& 0.06\\
B+H     & $c^{(\text{B})}$=1 $c^{(\text{H})}$=10	& {\bf 87.66}$^{\text{+++}}$	& 0.18\\\hline 
  \end{tabular}
  \caption{Full results on French dev set, for various tasks combinations, with and without stack propagation (H: nb of heads, D: nb of dependents, B: bag of labels, S: label multi-set). Col3-4: average LF, and standard deviation. $^{***}$: significative difference wrt first line ($p <$0.001). $^{\text{+++}}$: significative difference wrt B+H without stack propagation ($p <$0.001).}
  \label{tab-full-res-combis-taches-aux}
\end{center}
\end{table}

\end{document}